\PassOptionsToPackage{svgnames,dvipsnames}{xcolor}
\RequirePackage{caption}
\let\abovecaptionskip\undefined
\let\belowcaptionskip\undefined
\documentclass[10pt,journal,compsoc]{IEEEtran}

\newif\ifpeerreview
\peerreviewfalse

\newif\ifbiblatex
\biblatextrue

\def\papertitle{Monte Carlo Path Tracing and Statistical Event Detection for Event Camera Simulation}

\usepackage{amsmath,amssymb}
\usepackage{duckuments}
\newcommand{\DA}{\raisebox{0.2mm}{$\downarrow$}}
\newcommand{\UA}{\raisebox{0.2mm}{$\uparrow$}}

\usepackage[pdftex]{graphicx}
\graphicspath{{./fig/}}
\usepackage{subcaption}

\usepackage[most]{tcolorbox}
\usepackage{microtype}
\usepackage{comment}
\usepackage{confnames}

\usepackage{amsmath,amssymb}
\usepackage{mathtools}
\usepackage{upgreek}

\usepackage{color}
\usepackage{xcolor}

\usepackage{tabularray}
\UseTblrLibrary{booktabs,siunitx}
\newcommand{\tablefontsize}{\fontsize{7.0pt}{8.0pt}\selectfont}
\newcommand{\strong}[1]{\textcolor{MediumBlue}{\underline{\textbf{#1}}}}
\NewTableCommand{\BB}{\SetCell{bg=MistyRose}}

\makeatletter
\renewcommand\paragraph{\@startsection{paragraph}{4}{\z@}%
 {-0.8ex \@plus -0.3ex \@minus -0.3ex}%
 {-0.0em \@plus -0.2em \@minus -0.2em}%
 {\normalfont\normalsize\bfseries}}
\makeatother
\let\oldparagraph\paragraph
\renewcommand\paragraph[1]{\oldparagraph*{#1}}

\usepackage{enumitem}

\usepackage[linesnumbered,ruled]{algorithm2e}

\SetCommentSty{mycommfont}
\newcommand{\algfontsize}{\footnotesize}

\usepackage{cases}
\usepackage{physics2}
\usepackage{fixdif,derivative}
\usephysicsmodule{ab}

\usepackage{xparse}
\makeatletter
\NewDocumentCommand{\vb}{s}{\IfBooleanTF{#1}{\boldsymbol}{\mathbf}}
\newcommand\va[1]{\@ifstar{\vec{#1}}{\vec{\mathrm{#1}}}}
\newcommand\vu[1]{%
\@ifstar{\hat{\boldsymbol{#1}}}{\hat{\mathbf{#1}}}}
\makeatother

\usepackage{siunitx}
\DeclareSIUnit{\MB}{\mega\byte}
\DeclareSIUnit{\GB}{\giga\byte}
\DeclareSIUnit[quantity-product=]\percent{\char`\%}
\newcommand{\insertfig}{
 \centering
 \includegraphics[width=\linewidth]{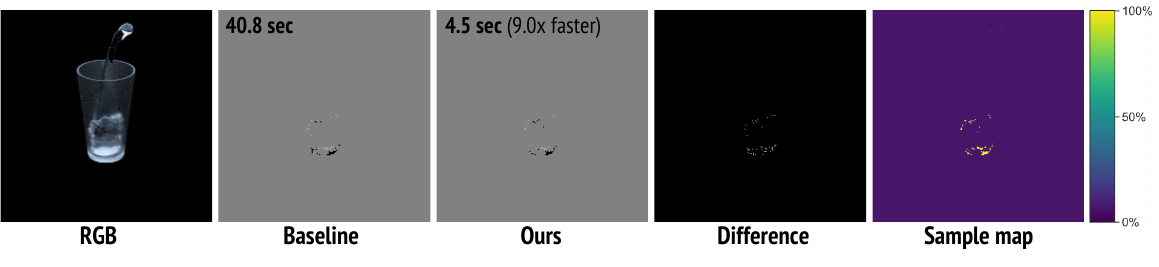}
 \vspace{-8mm}
 \captionof{figure}{The proposed method renders event videos using Monte Carlo path tracing via a statistical event detection approach. The proposed method achieves an order of magnitude faster than a simple approach of tracing a fixed number of path samples to each pixel. Specifically, the proposed method accelerates the computation time required to render a single frame from \qty{40.8}{\s} to \qty{4.5}{\s}, obtaining \num{9.0} times speed up. Meanwhile, the difference between our result and the baseline approach is sufficiently small, with the difference being observed at only a few pixels, and the sample map shows that the proposed method successfully eliminates path samples at the majority of pixels where events may not happen.}
 \label{fig:teaser}
 \vspace{3mm}
}

\makeatletter
\apptocmd{\@maketitle}{%
 \setcounter{figure}{0}%
 \centering\insertfig%
}{}{}
\makeatother

\ifbiblatex
\usepackage[
 bibstyle=ieee,
 citestyle=numeric,
 defernumbers=false,
 bibencoding=utf-8,
 sortcites=true,
 backend=biber,
 natbib=true,
 doi=true,
 url=true,
 isbn=false,
]{biblatex}

\fi

\usepackage[pdftex]{hyperref}
\hypersetup{
 colorlinks=true,
 urlcolor=magenta,
 citecolor=black,
 linkcolor=black,
 bookmarks=true,
 bookmarksnumbered=false,
 pdftitle={\papertitle{}},
}

\usepackage[switch]{lineno}

\newcommand{\paperID}{0002}

\title{\papertitle{}}

\author{%
 Yuichiro Manabe, Tatsuya Yatagawa~\IEEEmembership{Member,~IEEE}, \\
 Shigeo Morishima~\IEEEmembership{Member,~IEEE}, Hiroyuki Kubo~\IEEEmembership{Member,~IEEE}%
 \IEEEcompsocitemizethanks{
 \IEEEcompsocthanksitem
 Y. Manabe and H. Kubo are with the Graduate School of Science and Engineering, Chiba University, Chiba, 263--8522, Japan, \protect\\
 E-mail: 24wm3206@student.gs.chiba-u.jp (Y. Manabe), hkubo@chiba-u.jp (H. Kubo)
 \IEEEcompsocthanksitem
 T. Yatagawa is with the School of Social Data Science, Hitotsubashi University, Tokyo, 186--8601, Japan.\protect\\
 E-mail: tatsuya.yatagawa@r.hit-u.ac.jp
 \IEEEcompsocthanksitem
 S. Morishima is with the Graduate School of Science and Engineering, Waseda University, Tokyo, 162--0041, Japan.
 E-mail: shigeo@waseda.jp
 }
}

\usepackage[capitalize]{cleveref}

\newcommand{\crefplural}[1]{%
 \crefname{#1}{\csname #1name\endcsname}{\csname #1nameplural\endcsname}%
 \Crefname{#1}{\csname #1Name\endcsname}{\csname #1Nameplural\endcsname}%
 \csname #1nameplural\endcsname%
}

\usepackage{CJKutf8}

\usepackage{ragged2e}

\begin{document}

\IEEEtitleabstractindextext{%
  \begin{justify}
    \begin{abstract}
      This paper presents a novel event camera simulation system fully based on physically based Monte Carlo path tracing with adaptive path sampling. The adaptive sampling performed in the proposed method is based on a statistical technique, hypothesis testing for the hypothesis whether the difference of logarithmic luminances at two distant periods is significantly larger than a predefined event threshold. To this end, our rendering system collects logarithmic luminances rather than raw luminance in contrast to the conventional rendering system imitating conventional RGB cameras. Then, based on the central limit theorem, we reasonably assume that the distribution of the population mean of logarithmic luminance can be modeled as a normal distribution, allowing us to model the distribution of the difference of logarithmic luminance as a normal distribution. Then, using Student's t-test, we can test the hypothesis and determine whether to discard the null hypothesis for event non-occurrence. When we sample a sufficiently large number of path samples to satisfy the central limit theorem and obtain a clean set of events, our method achieves significant speed up compared to a simple approach of sampling paths uniformly at every pixel. To our knowledge, we are the first to simulate the behavior of event cameras in a physically accurate manner using an adaptive sampling technique in Monte Carlo path tracing, and we believe this study will contribute to the development of computer vision applications using event cameras.
    \end{abstract}
  \end{justify}
}

\maketitle
\thispagestyle{empty}

\IEEEraisesectionheading{
  \section{Introduction}
  \label{sec:intro}
}

\IEEEPARstart{T}{he} event camera equipped with event-based dynamic vision sensors (DVS) is an unconventional camera device to detect temporal brightness changes as ``events''~\cite{lichtsteiner2008,gallego2022event}. Different from conventional frame-based cameras, event cameras can detect events asynchronously and independently at each sensor pixel. One of the advantages of the event camera is its high temporal resolution and wide dynamic range, which are significantly higher than ordinary image sensors. Furthermore, the asynchronous and independent natures of event detection save on battery power consumption. These advantages of event cameras have been leveraged in various computer vision applications, including object tracking~\cite{zhu2017event, mitrokhin2018event}, 3D scanning~\cite{matsuda2015mc3d, zhou2018semi, takatani2021event}, optical flow estimation~\cite{benosman2014event, bardow2016simultaneous, zhu2018ev, tian2022event}, and simultaneous localization and mapping (SLAM)~\cite{kim2016real,vidal2018ultimate,zhou2021event}.

\begin{figure}[t!]
  \centering
  \includegraphics[width=\linewidth]{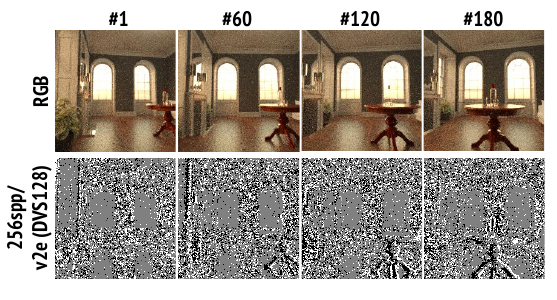}
  \caption{The result of a preliminary test of the v2e system~\cite{hu2021v2e} using the video frames obtained by Monte Carlo path tracing with 256 samples per pixel. Due to the lack of path samples, video frames input to the v2e system and the resulting event video are significantly noisy.}
  \label{fig:v2e-compare}
\end{figure}

As with other camera devices, simulating the behavior of event cameras is one of the most important research topics to facilitate the broader applications of the devices. Previously, there have been several studies conducted where simulators transform the input RGB video frames into a set of events~\cite{hu2021v2e, rebecq2018esim, joubert2021event}. There are also some application-specific simulators for object tracking~\cite{nehvi2021differentiable,zhu2021eventgan}, visual SLAM \cite{mueggler2017event}, and autonomous driving~\cite{gehrig2021dsec}. \Cref{tab:simulators} roughly compares previous simulators and the system proposed in this paper. As shown in this table, most existing simulators assume that input video frames are captured by traditional RGB cameras. However, they are unsuitable for simulating the behavior of event cameras provided that video frames obtained by physically based path tracing, which often contain much noise, are input to the simulators. As a preliminary study, we have tested the v2e system~\cite{hu2021v2e} using video frames obtained by Monte Carlo path tracing and showed the result in \cref{fig:v2e-compare}. In this figure, the input RGB video frames are calculated using 256 path samples per pixel. As shown in this figure, the noise level of the input video is not quite significant, while the noise is significantly amplified in the resulting event video. Thus, the physically accurate simulation of event cameras using Monte Carlo path tracing from synthetic 3D scenes have not been well explored.

One of the exceptions is the study by Tsuji et al.~\cite{tsuji2023event} that addresses the physically based rendering of event videos, where events are recorded at pixels with discrete periods. Their study focuses on applying denoising filters to generate noise-free video frames by saving on the computation budgets for tracing path samples. However, their study still does not offer any solution to an optimal distribution of path samples to obtain the event videos rather than obtaining clean video frames as the input for conventional event simulators. Another exception is ICNS Event-Based Camera Simulator (IEBCS)~\cite{joubert2021event}, which assumes a black-box rendering engine as with ESIM and processes the image sequence obtained by the renderer to simulate the behavior of event cameras. Unlike other simulators, IEBCS has utilized Blender Cycles, which is based on Monte Carlo path tracing, as a black-box rendering engine, and it can obtain ``physically plausible'' event videos from synthetic 3D scenes. However, adaptive sampling strategies exploiting the characteristics of event cameras have not yet been well investigated.

To address these problems, our study proposes a novel event camera simulation system, which directly performs physically accurate Monte Carlo path tracing to obtain a set of events. The proposed method first samples logarithmic luminance (i.e., brightness) from light sources and then performs stochastic event detection based on the sampled brightness. Specifically, we terminate path sampling early when we can statistically determine either that an event must happen or that no event will happen. We demonstrate the effectiveness of our method through visual, quantitative, and performance evaluations. Experimentally, our method is approximately ten times faster than simple Monte Carlo path tracing by appropriately eliminating the need for paths for pixels where events may not happen. Upon acceptance, the source code of our event rendering engine and 3D scene data will be released at \url{https://github.com/ichi-raven/MC-EBCS.git}.

\section{Related Work}
\label{sec:related-work}

\subsection{Dynamic vision sensor types}
\label{ssec:dvs-types}

During the development of event cameras, several sensors have been introduced. Currently, most off-the-shelf event cameras are equipped with one of three sensors~\cite{gallego2022event}, namely either a dynamic vision sensor (DVS), an asynchronous time-based image sensor (ATIS), or a dynamic and active pixel vision sensor (DAVIS). The DVS~\cite{lichtsteiner2008}, apart from conventional frame-based cameras, offered an asynchronous and per-pixel response to brightness changes. This capability imitates the adaptive and spiking nature of biological responses and provides a stream of variable-rate events that each encode a change in logarithmic illuminance at precise temporal resolutions. Advancing the capabilities of DVS, ATIS~\cite{posch2011qvga} incorporates a dual-subpixel architecture; one is for detecting brightness changes, and the other is for measuring absolute intensities. The letter subpixel is used to control the reset level of the brightness change detector, and thereby, ATIS provides a significantly high dynamic range of more than \qty{120}{\dB}, enabling more sensitive acquisition of scene details across broad luminance ranges. Similar to ATIS, DAVIS~\cite{brandli2014} is also equipped with active pixel sensors (APS) to detect the absolute illuminance of the scene. DAVIS uses the APS to acquire ordinary video frames rather than to control the reset level. Therefore, the dynamic range of DAVIS is lower than that of ATIS, but it is capable of acquiring dynamic event data alongside static intensity frames.

In this paper, we focus on the nature of these sensors for detecting brightness changes, i.e., the fundamental behavior of DVS. However, as our system is based on Monte Carlo path tracing, which is normally used to obtain RGB video frames, it may also be possible to enable our system to simulate the behavior of ATIS and DAVIS by adding subtle implementation.

\begin{table}[t!]

  \centering
  \caption{Comparison of event camera simulators}
  \label{tab:simulators}
  {\tablefontsize
    \begin{tblr}{colspec={X[l]|Q[l]Q[l]Q[c]},rowsep=0.8mm}
      \toprule
                                    & Inputs                   & Rendering Engine         & Adaptive     \\
      \midrule
      ESIM~\cite{rebecq2018esim}    & Video, \strong{3D scene} & Rasterize                & No           \\
      v2e~\cite{hu2021v2e}          & Video                    & N/A (RGB camera)         & N/A          \\
      IEBCS~\cite{joubert2021event} & Video, \strong{3D scene} & \strong{MC path tracing} & No           \\
      ESIM-AD~\cite{tsuji2023event} & (Noisy) Frames           & N/A (pre-rendered)       & Yes          \\
      \midrule[dotted]
      Ours                          & \strong{3D scene}        & \strong{MC path tracing} & \strong{Yes} \\
      \bottomrule
    \end{tblr}}

\end{table}

\subsection{Event camera simulators}
\label{ssec:ev-simulators}

When we develop or test the applications, it will be useful to obtain the event video from RGB video frames with a typical frame rate (e.g., 30--60 frames per second). To this end, previous studies have solved two problems, i.e., increasing the frame rate of an input video and transforming the video frames into a set of events. Based on the behavior of DVS, a pioneering work by Katz et al.~\cite{katz2012live} used a \qty{200}{\Hz} frame rate camera and synthesized DVS events with a \qty{5}{\ms} time resolution. Rebecq et al. proposed a system called ``ESIM''~\cite{rebecq2018esim}, which obtains events at arbitrary time points from synthetic 3D scenes using a black-box rendering engine. However, these systems would be impractical when applying Monte Carlo path tracing to the rendering engine because obtaining noise-free video frames with Monte Carlo path tracing requires considerable computation time. Recently, Hu et al. proposed another system called ``v2e''~\cite{hu2021v2e}, where the input video frames are increased in frame rate using a deep-learning-based method, i.e., Super SloMo~\cite{jiang2018super}, and then the high frame rate video is converted into a set of events by carefully considering the behavior of real DVS sensors. However, the v2e system also requires noise-free video frames, and we need significant computation time when we input video obtained by MC path tracing to solve the rendering equation~\cite{kajiya1986rendering}. Recently, Joubert et al. proposed the IEBCS system, another event camera simulator using a black-box rendering engine~\cite{joubert2021event}. Although they demonstrated that their system can employ Blender Cycles, a rendering engine based on Monte Carlo path tracing, their system still does not address the problem of high computational costs to obtain clean video frames through rendering. The study by Tsuji et al.~\cite{tsuji2023event} also addresses the physically based rendering of event videos. However, their research has focused only on denoising filters to generate event videos efficiently rather than improving the performance of Monte Carlo path tracing.

Thus, despite the continuous efforts to develop event camera simulators, no specialized approach to Monte Carlo path tracing for rendering event videos has been well investigated yet. To address this issue, we propose a novel event camera simulation system that traces path samples adaptively for pixels where events are more likely to happen.

\subsection{Applications of event cameras}
\label{ssec:ebc-applications}

Here, we introduce several key applications of event cameras, particularly in the field of computer vision. Owing to the high temporal resolution and low latency response of event cameras, they are particularly well-suited for object tracing, where event cameras can track fast-moving objects accurately without the need for high-speed cameras~\cite{zhu2017event,mitrokhin2018event}. Optical flow estimation is another area where event cameras can be significant. Traditional methods often struggle with fast-moving objects or low-light conditions, where the lack of sufficient temporal resolution and sensitivity makes the estimation difficult. To address this problem, several studies have proposed accurate optical flow estimation systems under demanding conditions owing to their capability of asynchronously capturing subtle changes in the scene with high temporal precision~\cite{benosman2014event,zhu2018ev,tian2022event}. Furthermore, event cameras are applied for 3D scene recognition, including the techniques for 3D scanning~\cite{matsuda2015mc3d,zhou2018semi,takatani2021event} and visual SLAM~\cite{kim2016real,vidal2018ultimate,zhou2021event}. As outlined, the applications of event cameras can potentially solve problems in the field of computer vision from a different angle, addressing the limitations of traditional methods. For numerous other studies and applications, we refer the reader to a comprehensive survey on the event-based vision~\cite{gallego2022event}.

\begin{algorithm}[t!]
  \caption{Pseudocode of the proposed method}
  \label{alg:pseudocode}
  {\algfontsize
  \KwData{The number of path samples $N_{\text{spp}}$, the total number of frames $N_{\text{frame}}$, and event threshold $\theta$}
  \KwResult{Set of events $\vb{E}$}
  \DontPrintSemicolon
  \SetNoFillComment
  \BlankLine
  \For{each pixel $\vb{q}$}{
  \tcc*[l]{Simply rendering the 1st frame}
  $\vb{B} \leftarrow \text{\texttt{tracePaths}}(\vb{q}, 1, N_{\text{spp}})$ \;
  \tcc*[l]{Detect events for the rest of frames}
  $\vb{E} = \{ \Phi \}$ \;
  $\{ \mu_a, \sigma_a^2 \} \leftarrow \text{\texttt{calcMeanVar}}(\vb{B})$ \;
  \For{$s = 2, \ldots, N_{\text{frame}}$}{

  $\vb{B} \leftarrow \text{\texttt{tracePaths}}(\vb{q}, s, 256)$ \;
  $N_{\text{total}} \leftarrow 256$ \;
  $N_{\text{batch}} \leftarrow \min \ab\{ 64, |N_{\text{spp}} - 256|\}$ \;
  \tcc*[l]{Incremental path sampling}
  \While{$N_{\text{\normalfont total}} < N_{\text{\normalfont spp}}$}{
    $\vb{B} \leftarrow \vb{B} \cup \text{\texttt{tracePaths}}(\vb{q}, s, N_{\text{batch}})$ \;
    $\{ \mu_b, \sigma_b^2 \} \leftarrow \text{\texttt{calcMeanVar}}(\vb{B})$ \;
    \BlankLine
    \tcc*[l]{One-tailed Student's $t$-test}
    $t = \tfrac{|\mu_b - \mu_a| - \theta}{(\sigma_a^2 + \sigma_b^2) / \sqrt{N_{\text{total}}}}$ \tcp*[r]{\cref{eq:t-value}}
    \If{$P(t) < \alpha$}{
      break \;
    }
    $N_{\text{total}} \leftarrow N_{\text{total}} + N_{\text{batch}}$ \;
    $N_{\text{batch}} = 64$ \;
  }
  \tcc*[l]{Event detection}
  \If{$|\mu_b - \mu_a| > \theta$}{
    $p \leftarrow \text{sign}(\mu_b - \mu_a)$ \;
    $\vb{E} \leftarrow \vb{E} \cup \{ (\vb{q}, t, p) \}$ \;
    \tcc*[l]{Update statistics}
    $\{ \mu_a, \sigma_a^2 \} \leftarrow \{ \mu_b, \sigma_b^2 \}$
  }
  }
  }}
\end{algorithm}

\section{Physically Based Event Renderer}
\label{sec:method}

The basic principle of event detection by DVS or other event sensors is that an event is triggered when the brightness change $| \Delta B |$ reaches a temporal contrast threshold $\theta \geq 0$. More specifically, the brightness here is the logarithmic illuminance at each sensor pixel. Assume that the last event is detected time $\Delta t$ earlier than $s_l$, then an event is triggered when the following equation holds.
\begin{equation}
  \Delta B(\vb{q}_k, s_l) = B(\vb{q}_k, s_l) - B(\vb{q}_k, s_l - \Delta t) = p_k C,
  \label{eq:event-detection}
\end{equation}
where $\vb{q}_k$ and $s_l$ are the pixel location and time period where an event occurs, and $p_k$ is the polarization of an event and takes either $+1$ or $-1$. Here, the brightness $B$ is involved with the logarithm of photocurrent $I$ detected as the image sensor, such as CMOS. Therefore, considering simply, we can obtain a set of events from animated 3D scenes by rendering frames at discrete time periods and then apply \cref{eq:event-detection} to transform the video frames into a set of events. However, such a simple approach is not intractable with Monte Carlo path tracing because it requires noise-free video frames and Monte Carlo path tracing typically requires huge computational time to obtain such noise-free video frames.

What is worth noting here is that events often occur at only a few pixels of each video frame. Therefore, tracing a predefined number of path samples to every pixel is not efficient. The proposed method aims to solve this problem by adaptively sampling path samples to pixels where events are more likely to happen. To this end, we sample logarithmic brightness from light sources and then perform statistical event detection based on hypothesis testing. Specifically, we can reasonably assume that the distributions of pixel brightness values follow a normal distribution, which is justified by the central limit theorem, and the distribution of the gap of pixel brightness values also follows a normal distribution as well. Then, we perform Student's $t$-test to test the hypothesis whether we can discard the null hypothesis for event non-occurrence.

As an overview, we provided a pseudocode in \cref{alg:pseudocode}. In this pseudocode, the subroutine \texttt{tracePaths}$(\vb{q}, s, N)$ performs tracing $N$ paths at pixel $\vb{q}$ at time $s$ and obtains a set of brightness values for sampled paths. Another subroutine \texttt{calcMeanVar}$(\vb{B})$ calculates the mean and the variance of brightness values in the set $\vb{B}$.

\subsection{Sampling logarithmic luminances}
\label{sec:sample-logarithmic}

To compute the probability density function for the logarithmic luminance at each pixel, we first need to sample the logarithmic luminance from light sources rather than the luminance itself. In this section, we explain how we can sample the logarithmic luminance using Monte Carlo path tracing. In ordinary Monte Carlo path tracing, the luminance in units of \unit{\W\per\steradian\per\square\m} is carried by a path sample, and the luminance samples are integrated over the hemispherical region of the camera sensor. Then, the irradiance in units of \unit{\W\per\square\m} can be obtained for each sensor pixel. Finally, the intensity of the irradiance will be electronically converted to photocurrent $I$ at the image sensor at each pixel. Therefore, ordinary Monte Carlo path tracing solves the following integration at each pixel to obtain how much intensity each pixel will receive.
\begin{equation}
  L = \mathbb{E}_{\mathbf{x} \sim \mathcal{X}} \ab[ \frac{f(\vb{x})}{p(\vb{x})} ],
  \label{eq:sample-mean}
\end{equation}
where $\vb{x}_i \in \mathcal{X}$ is a path sample, which is sampled from the space of traveling paths of lights $\mathcal{X}$ by the probability of $p(\vb{x}_i)$, $f$ is the integrand (here, the contribution corresponding to each path is calculated) and $N_{\text{spp}}$ is the total number of path samples collected to obtain an approximate solution to the integral. While this is the probabilistic integration scheme, we can calculate the variance of the sample mean in \cref{eq:sample-mean} as well.

In contrast to the conventional approach, we collect logarithmic luminances rather than luminances themselves. In this manner, we can calculate the distribution of logarithmic luminance values. Specifically, \cref{eq:sample-mean} will be transformed into the following form.
\begin{equation}
  \ln L = \mathbb{E}_{\mathbf{x} \sim \mathcal{X}} \ab[ \frac{\ln(f(\vb{x}))}{p(\vb{x})} ].
  \label{eq:log-sample-mean}
\end{equation}
Then, we can calculate $\mathbb{V}\ab[\ln L]$, the variance of logarithmic luminance $\ln L$, in the following form.
\begin{equation}
  \mathbb{V}\ab[\ln L] = \mathbb{E}_{\vb{x} \sim \mathcal{X}} \ab[ \frac{\ab(\ln(f(\vb{x}_i)) - \ln L)^2}{p(\vb{x}_i)} ].
  \label{eq:log-sample-variance}
\end{equation}
Using \cref{eq:log-sample-mean,eq:log-sample-variance}, we can define a normal distribution for the logarithmic luminance at each pixel. We assume here that the central limit theorem can apply to the above case and when the number of path samples used to discretely calculate the sample mean in \cref{eq:log-sample-mean} and the sample variance \cref{eq:log-sample-variance} is sufficiently large, we can reasonably assume that the resulting distribution of logarithmic luminances follows the normal distribution.

\paragraph{Discussion}

We sample logarithmic luminance values rather than raw luminance values to calculate the sample mean and sample variance of the logarithmic luminance. As the logarithmic function is concave, Jensen's inequality gives the following relationship about the sample mean.
\begin{equation}
  \ln \mathbb{E}\ab[L] \geq \mathbb{E}\ab[\ln L].
  \label{eq:jensen}
\end{equation}
Our method samples the logarithm of luminance levels to calculate their variance, rather than directly calculating the logarithm of the sample mean of luminance levels. However, whether we first take the logarithm of the sample mean or the sample mean of the logarithm may lead to different results. To our knowledge, in the real circuit of a DVS sensor, the photocurrent is induced by photons hitting the photoreceptor, and the photocurrent is converted to the logarithmic value by a logarithmic circuit. In this process, both the photocurrent and the logarithmic value are integrated over time, and it is hard to determine which is more dominant. If the time integration of the photocurrent would be more dominant than that of the logarithmic value, \cref{eq:jensen} implies that our approach results in both the sample mean and variance being lower than those obtained through the logarithm of the sample mean method. However, during our experiments, we did not find any problems that may be induced by taking solely the average of the logarithmic luminance.

\begin{figure}[t!]
  \centering
  \includegraphics[width=\linewidth]{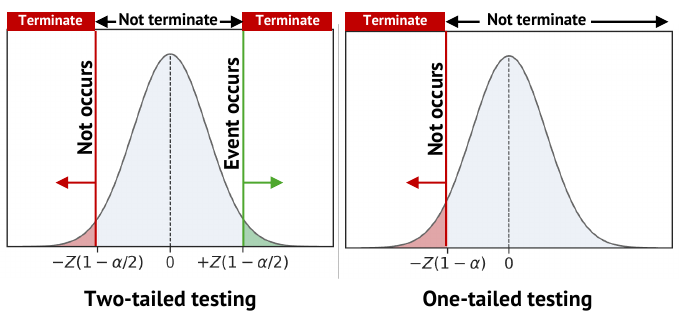}
  \caption{Illustration representing the difference between one-tailed and two-tailed hypothesis testing. While two-tailed testing discards the null hypothesis in both cases where an event occurs or no event occurs, one-tailed testing discards the hypothesis only in the case where no event occurs.}
  \label{fig:t-test}
\end{figure}

\subsection{Statistical event detection by hypothesis testing}
\label{ssec:hypothesis-testing}

As we mentioned previously, the pixels where any event occurs are highly sparse. Therefore, we can skip tracing path samples at most pixels where events may not happen. Given the normal distributions for the logarithmic luminance values of a pixel at two distant time points, we perform hypothesis testing on the difference between the sample means of the logarithmic luminance at each time point.

Let us assume that we now have the normal distributions $\mathcal{N}(\mu_a, \sigma_a^2)$ and $\mathcal{N}(\mu_b, \sigma_b^2)$, representing the distributions of logarithmic luminance values at two time points $a$ and $b$. Assuming that the distributions of logarithmic luminances at different time points are independent, we can calculate the normal distribution for the possible gap between the luminance values at these two distant time periods.
\begin{equation}
  \Delta \ln L \sim \mathcal{N}\ab(\ab| \mu_b - \mu_a |, \sigma_a^2 + \sigma_b^2).
\end{equation}
Using this normal distribution for the gap of logarithmic luminance at two distant time points, the Student's t-test applies to determine whether the population mean of the gap is less than or equal to event threshold $\theta$ or not. Thus, the null hypothesis of this test will be as follows.
\vspace{0.2\baselineskip}
\begin{tcolorbox}[colback=white,boxrule=0.3mm]
  \textbf{Null hypothesis:} \\[1mm]
  The population mean of the brightness gap is equal to the event threshold $\theta$.
\end{tcolorbox}
\vspace{0.2\baselineskip}
To test this hypothesis using Student's $t$-test, we calculate the following $t$-value as
\begin{equation}
  t = \frac{\ab| \mu_b - \mu_a | - \theta}{\sqrt{(\sigma_a^2 + \sigma_b^2) / N}},
  \label{eq:t-value}
\end{equation}
where $N$ is the number of samples used to calculate the sample mean and the sample variance (i.e., the number of path samples in the context of Monte Carlo path tracing) at the time point $b$. It is worth noting that we do not test whether the population mean of the logarithmic luminances at two time periods are equal or not, which can be tested by Welch's $t$-test. Rather than that, we consider the distribution of the population means of the brightness gap and test whether the gap is equal to the event threshold not by Student's $t$-test.

When the null hypothesis is discarded in this testing, there are two possible options. One is the population mean of the gap is significantly \textit{larger} than $\theta$, and the other is the population mean of the gap is significantly \textit{smaller} than $\theta$. In our implementation, as the difference is shown in \cref{fig:t-test}, we empirically employ the \textit{one-tailed} test to check if the population mean of the gap is significantly smaller than the threshold. To ensure \qty{95}{\%} confidence interval, we set the significance level $\alpha = 0.05$. When $P(t) < \alpha$, where $P(t)$ is the $p$-value of $t$, the null hypothesis is discarded, and we can statistically ensure that no event will be detected even by increasing the sample numbers anymore, and we can terminate path sampling before reaching the sample number to the predefined one.

\begin{figure}[t!]
  \centering
  \includegraphics[width=\linewidth]{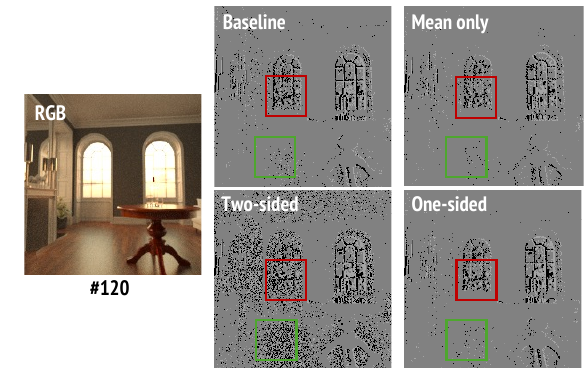}
  \caption{Comparison of event rendering results obtained by mean-only, two-tailed, and one-tailed approaches. While the result of the two-tailed approach includes significantly much noise, the mean-only and one-tailed approaches obtain equivalent results.}
  \label{fig:ablation-study}
\end{figure}

\paragraph{Discussion}

To check whether an event happens at a specific frame, we may have several options other than the one-tailed hypothesis testing that we employed. The simplest one is to consider only the difference of sample means at two distant periods. Another choice is the two-tailed hypothesis testing of discarding the null hypothesis when the brightness gap is likely more than the event threshold. \Cref{fig:ablation-study} compares the event rendering results of our one-tailed approach with mean-only and two-tailed approaches. Comparing their results with the baseline approach, which we describe in detail in \cref{sec:results}, the one obtained by the two-tailed approach is significantly noisy. This is because the estimate for the brightness at a time period where an event occurs is used to check whether another event occurs at a later time period. When the null hypothesis is discarded and path sampling is terminated early at the time period of event occurrence, the estimate for the brightness gap can be inaccurate. Therefore, we need to trace a sufficient number of paths for the pixel where an event occurs, and thus, we employed one-tailed hypothesis testing. As the results of one-tailed and mean-only approaches are visually similar, we will elaborate on the analysis later in \cref{sec:results}.

\subsection{Implementation details}
\label{ssec:implementation}

The techniques shown above are implemented as a rendering system with C/C++, which is built upon pbrt-v4~\cite{pbrt-v4}, an open-source physically based rendering engine based on Monte Carlo path tracing. Although we implemented our method by manipulating \texttt{PathIntegrator} that performs ordinary unidirectional path tracing, our method does not depend on which path tracing method is used. Therefore, we can implement the same algorithm over other types of path tracing methods, such as bidirectional path tracing~\cite{veach1995bidirectional} and photon mapping~\cite{jensen1996global,hachisuka2009stochastic,qin2015unbiased}. Furthermore, the proposed method can be implemented on top of other Monte Carlo path tracing engines and is not limited to the implementation over pbrt-v4.

In practice, we can assume that the central limit theorem can be considered satisfied and the distribution of sample mean can be approximated as a normal distribution. In our current implementation, we conservatively perform the first hypothesis testing when we have traced 256 samples per pixel. Subsequently, we perform hypothesis testing at every pixel and we terminate sampling paths at pixels where the null hypothesis has been discarded. For the remaining pixels, we add 64 more samples and perform hypothesis testing again. Unless the null hypothesis is not discarded, we continue path samples until the number of path samples for a pixel reaches the predefined number of maximum samples (we currently use 4096 samples).

During this process, since we employ the one-tailed t-test to discard the null hypothesis of the non-occurrence of any events, we result in tracing a maximum number of paths for the pixels where an event occurs. On the one hand, such behavior may unnecessarily increase the total number of path samples required. As shown in \cref{fig:ablation-study}, two-tailed hypothesis testing where both the hypotheses of event occurrence and non-occurrence result in noisy event video. On the other hand, one-tailed hypothesis testing can prevent such unexpected events from being detected, and all the following events can deviate in timing from the time point they should occur. However, even this approach, which is not perfectly optimized, can obtain approximately 10 times the speed up compared to a simple approach, and furthermore, the resulting event video is sufficiently similar to the one obtained by the simple approach. Thus, our current choice of one-tailed hypothesis testing is reasonable in practice.

\begin{figure*}[!p]
  \centering
  \begin{tblr}{
      colspec={X[c]},
      colsep=0mm,
      rowsep=3mm,
    }
    \begin{minipage}[t]{0.48\linewidth}
      \setcounter{subfigure}{0}
      \centering
      \includegraphics[width=\linewidth]{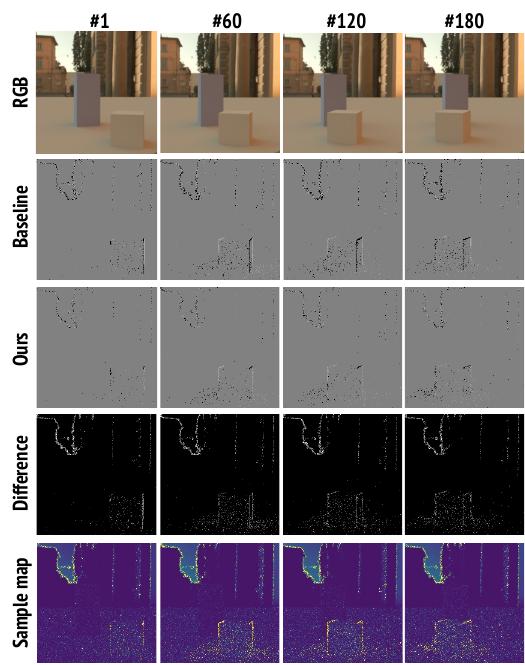}
      \subcaption{The ``two boxes'' scene (threshold: $\theta=0.5$)}
      \label{fig:result-tb}
    \end{minipage}
    \hfill
    \begin{minipage}[t]{0.48\linewidth}
      \centering
      \includegraphics[width=\linewidth]{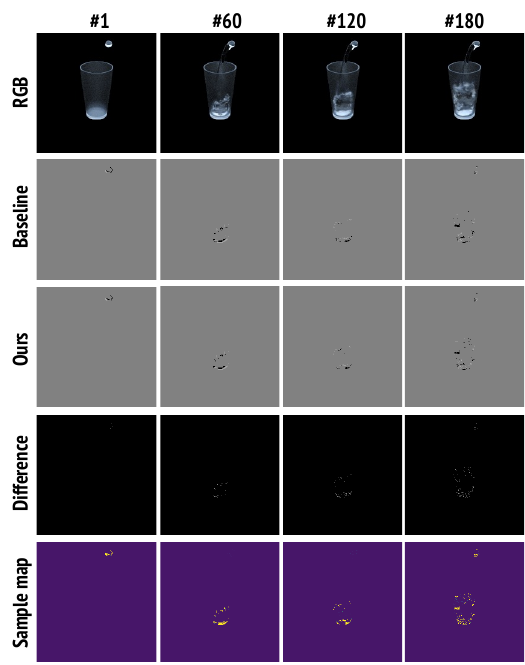}
      \subcaption{The ``cup of water'' scene (threshold: $\theta=0.5$)}
      \label{fig:result-cw}
    \end{minipage} \\
    \begin{minipage}[t]{0.48\linewidth}
      \centering
      \includegraphics[width=\linewidth]{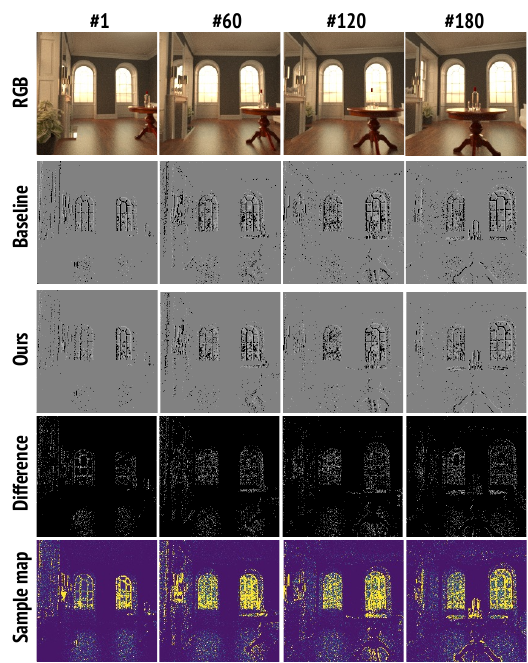}
      \subcaption{The ``living room'' scene (threshold: $\theta=3.0$)}
      \label{fig:result-lr}
    \end{minipage}
    \hfill
    \begin{minipage}[t]{0.48\linewidth}
      \centering
      \includegraphics[width=\linewidth]{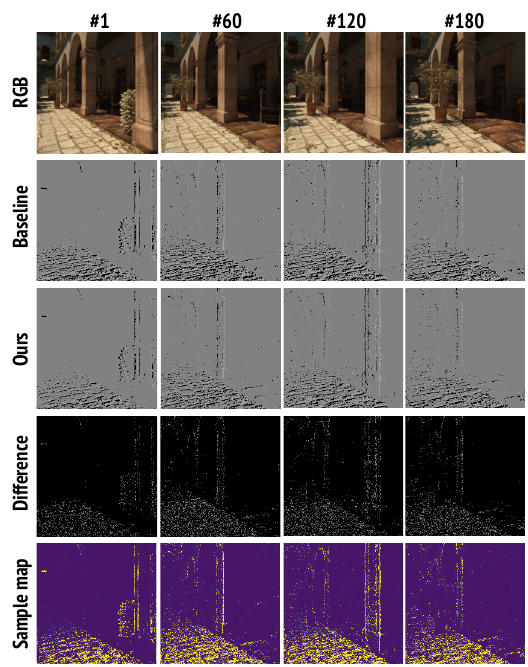}
      \subcaption{The ``San Miguel'' scene (threshold: $\theta=0.6$)}
      \label{fig:result-san}
    \end{minipage}
  \end{tblr}
  \caption{Rendering results of the proposed system and the sampling ratio compared to the uniform sampling. The color bar applied to the sample maps is the same as that in \cref{fig:teaser}.}
  \label{fig:render-results}
\end{figure*}

\section{Results and Discussion}
\label{sec:results}

To demonstrate the performance of our rendering system for event videos, we conducted experiments across various scenes, including the ``two boxes,'' ``cup of water,'' ``living room,'' and ``San Miguel'' scenes. The ``two boxes'' and ``cup of water'' scenes are synthesized by ourselves using Blender, the open-source content creation software, and the ``living room'' and ``San Miguel" scenes are prepared by manipulating the scenes files, which are provided publicly on the Internet~\cite{pbrt-scenes,bitterli16pbrt}, such that the camera moves through the scenes. In the following experiments, our rendering system is tested on a computer equipped with an Intel Core i9-13900K CPU (\qty{5.80}{\GHz}, 24 cores), NVIDIA RTX A6000 GPU, and \qty{128}{\GB} of RAM. Each rendering result is rendered using 4096 path samples for each pixel and consists of 240 discrete time steps. Each frame of the resulting event videos has \numproduct{200x200} pixels, and the total number of pixels we need to process is approximately ten million pixels.

The baseline method that we compare with the proposed method is a simple approach of tracing a maximum number of path samples (i.e., 4096 path samples) for every pixel. Therefore, in the experimental results below, we can say that the more similar the results of our method are to those of the baseline method, the better the results of our method are. In this baseline method, we do not consider some precise behavior of real event-cameras, such as motion blur, current leakage, and random threshold variation because considering these behaviors makes the quantitative evaluation of the proposed method in comparison with this baseline approach difficult.

\subsection{Visual assessment on rendering results}
\label{ssec:qualitative}

\Cref{fig:render-results} shows the event rendering results for four sample scenes, where the rendering results of baseline and our approaches are shown in the second and third row, the difference between them is shown in the fourth row, and the ratio of the number of path sample of ours to that of the baseline method at each pixel is visualized in the bottom row. Due to the granular nature of event videos, it is somewhat difficult to judge whether the results of the baseline and our methods are sufficiently similar or not; we can say that there are no significant differences in terms of detecting events at locations close to each other.

In contrast, we also find that the proposed method overlooks some events that have been detected by the baseline method, and the number of events detected for each frame is somewhat smaller in the proposed method than in the baseline method. Such omitted events are considered due to the error of the first kind of hypothesis testing, where the null hypothesis is discarded to result in no event occurrence even though whether an event occurs is still uncertain. To elaborate on the impact of such omitted events on the event rendering performance, we have conducted quantitative evaluations for the proposed method, which we will describe in the next subsection.

\subsection{Quantitative evaluations}
\label{ssec:quantitative}

For quantitative evaluation of the proposed method, we have employed several metrics, i.e., root-mean-squared error (RMSE), peak signal-to-noise ratio (PSNR), F1 score, and chamfer distance, following the prior study by Tsuji et al.~\cite{tsuji2023event}. RMSE and PSNR are calculated for event video frames in which negative, null, and positive events are stored as pixel values of 0.0, 0.5, and 1.0, respectively. F1 score and chamfer distance are the ones for point clouds~\cite{tankstemples} but modified to be aware of the difference of polarization of events~\cite{tsuji2023event}. To calculate these two metrics, a set of events is converted to ``signed'' 3D point clouds where each point is represented by 3D coordinates $(q_x, q_y, t)$ with polarization sign $p$. As event video records collections of events being detected at spatiotemporal points, assessing them solely based on image quality metrics would inaccurately penalize minor discrepancies in event detection timing. Therefore, we use these four metrics to provide a more comprehensive evaluation that considers the spatiotemporal characteristics of event detection results.

\Cref{tab:quantitative} shows the values of these evaluation metrics for the proposed method using one-tailed hypothesis testing and an alternative approach of considering only the difference of means. In this table, cells are highlighted by red background, when the score in the cell is better comparing mean-only and our approaches. Although, due to the granular nature of event videos, pure image-based metrics, such as RMSE and PSNR, are not absolutely favorable, the results of the spatiotemporal analysis with F1 score and chamfer distance imply the high quality of event videos obtained by our method. Furthermore, the scores achieved by our method are significantly better than those of the mean-only approach, showing that hypothesis testing is effective in determining the timing for path sampling termination. Furthermore, the proposed method achieves six or more times speed-up compared to the baseline approach. With these results of high evaluation scores and time performance, we can conclude that the proposed method has offered a reasonable judgment for when to terminate the path sampling by retaining high-quality output event videos.

\begin{table}[t!]
  \centering
  \caption{Quantitative evaluation of the proposed method}
  \label{tab:quantitative}
  {\tablefontsize
    \begin{tblr}{
      colsep=2mm,
      colspec={X[l]Q[l]|X[l]X[l]X[l]X[l]},
      row{1}={m},
      row{2,7,8,13,14}={cmd={\sisetup{round-precision=1,round-mode=places}}},
      row{3,4,9,10}={cmd={\sisetup{round-precision=2,round-mode=places}}},
      row{5,11}={cmd={\sisetup{round-precision=3,round-mode=places}}},
      row{6,12}={guard},
        }
      \toprule
                                 &                 & {{{Two boxes}}}                 & {{{Cup of water}}}              & {{{Living room}}}               & {{{San Miguel}}}                \\
      \midrule
      Baseline                   & Time            & \qty{39.51871068}{\s}           & \qty{40.80625988}{\s}           & \qty{144.2223827}{\s}           & \qty{161.9354203}{\s}           \\
      \midrule[0.1mm]
      \SetCell[r=6]{l} Mean only & RMSE\,$\DA$     & \BB \num{0.088391098}           & \num{0.018197608}               & \num{0.135730824}               & \num{0.127292745}               \\
                                 & PSNR\,$\UA$     & \BB \num{20.88226}              & \num{34.84537}                  & \num{17.28777}                  & \num{17.94361}                  \\
                                 & F1 score\,$\UA$ & \num{0.82506}                   & \num{0.98208}                   & \num{0.9753}                    & \num{0.98911}                   \\
                                 & PSCD\,$\DA$     & \num{148e-5}                    & \num{11.7e-5}                   & \num{10.8e-5}                   & \num{7.2e-5}                    \\
      \cmidrule[dotted]{1-Z}
                                 & Time\,$\DA$     & \qty{2.664112446}{\s}           & \qty{3.756094004}{\s}           & \qty{21.557634}{\s}             & \qty{22.01685693}{\s}           \\
                                 & Speed up\,$\UA$ & \num{14.83372473}{{{$\times$}}} & \num{10.86401454}{{{$\times$}}} & \num{6.69008402}{{{$\times$}}}  & \num{7.355065296}{{{$\times$}}} \\
      \midrule[0.1mm]
      \SetCell[r=6]{l} Ours      & RMSE\,$\DA$     & \num{0.091326039}               & \BB \num{0.017756824}           & \BB \num{0.134046902}           & \BB \num{0.125633765}           \\
                                 & PSNR\,$\UA$     & \num{20.42587}                  & \BB \num{35.05691}              & \BB \num{17.36174}              & \BB \num{18.05556}              \\
                                 & F1 score\,$\UA$ & \BB \num{0.90364}               & \BB \num{0.98356}               & \BB \num{0.97989}               & \BB \num{0.9898}                \\
                                 & PSCD\,$\DA$     & \BB \num{77.9e-5}               & \BB \num{11.5e-5}               & \BB \num{9.3e-5}                & \BB \num{7.0e-5}                \\
      \cmidrule[dotted]{1-Z}
                                 & Time\,$\DA$     & \qty{4.82984385}{\s}            & \qty{4.546246542}{\s}           & \qty{22.59764413}{\s}           & \qty{22.22948018}{\s}           \\
                                 & Speed up\,$\UA$ & \num{8.182192201}{{{$\times$}}} & \num{8.975813234}{{{$\times$}}} & \num{6.382186651}{{{$\times$}}} & \num{7.284714667}{{{$\times$}}} \\
      \bottomrule
    \end{tblr}
  }
\end{table}

\begin{table}[t!]
  \centering
  \caption{The speed-up ratios for scenes rendered with various image sizes}
  \label{tab:speed-up}
  {\tablefontsize
    \begin{tblr}{
        colspec={
            X[c] |
            *{4}{X[c,si={table-format=2.1,round-mode=places,round-precision=1}]}
          },
        row{1}={guard},
        colsep=1mm,
      }
      \toprule
      Size                   & Two boxes              & Cup of water            & Living room            & San Miguel              \\
      \midrule
      \numproduct{100x100}   & 8.684397{{{$\times$}}} & 15.396928{{{$\times$}}} & 7.895181{{{$\times$}}} & 10.014389{{{$\times$}}} \\
      \numproduct{200x200}   & 8.771519{{{$\times$}}} & 14.746755{{{$\times$}}} & 8.281566{{{$\times$}}} & 7.174531{{{$\times$}}}  \\
      \numproduct{400x400}   & 9.638422{{{$\times$}}} & 15.961017{{{$\times$}}} & 7.873840{{{$\times$}}} & 5.647452{{{$\times$}}}  \\
      \numproduct{800x800}   & 8.956468{{{$\times$}}} & 15.859065{{{$\times$}}} & 7.547483{{{$\times$}}} & 4.908301{{{$\times$}}}  \\
      \numproduct{1600x1600} & 9.949298{{{$\times$}}} & 15.539180{{{$\times$}}} & 7.463219{{{$\times$}}} & 4.431800{{{$\times$}}}  \\
      \bottomrule
    \end{tblr}}
\end{table}

\subsection{Time performance improvement}
\label{ssec:performance}

We have measured the total computational times and time performance improvements for \numproduct{200x200} video frames in the above experiments. Since the time performance improvement may be affected by the frame size, we have examined the time performance with video frames with varying sizes. \Cref{tab:speed-up} shows the performance improvement obtained by our method to four test scenes with various frame sizes. As shown by this table, the degree of time performance improvement is mainly affected by the type of scene, and the degree of improvement is consistent despite the change in the video frame size.

In contrast, the performance improvement for the San Miguel scene decreases as the size of the video frame increases. This is considered due to a comparatively large number of pixels where an event occurs. Our method, which employs one-tailed testing, traces as many as $N_{\text{spp}}$ paths for pixels where an event occurs, and thus, the performance improvement is largely dependent on the number of pixels where an event does not occur and the null hypothesis is discarded as early as possible. For the San Miguel scene, due to the presence of many events, the linear increase of the computational time for pixels with events affects more dominantly the computational time than the performance improvement gained at the pixels where the null hypothesis is discarded.

\begin{figure}[t!]
  \centering
  \includegraphics[width=\linewidth]{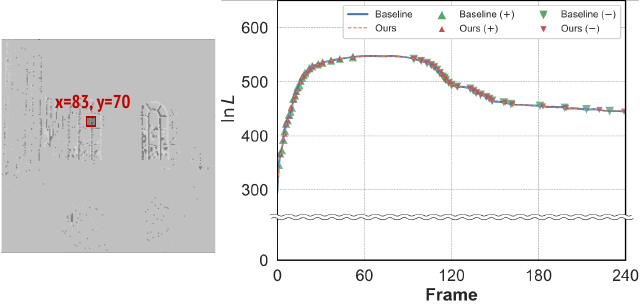}
  \caption{A detailed analysis of the difference in event occurrence between the baseline and our method. The change in $\ln L$ at pixel $(83, 70)$, which exhibits the maximum event deviation error in the ``living room'' scene, is shown with the time points of event occurrence, where positive and negative events are marked by ``$\blacktriangle$'' and ``$\blacktriangledown$'' symbols, respectively. Best viewed on screen.}
  \label{fig:max-error}
\end{figure}

\subsection{Analysis on event occurrence discrepancy}
\label{ssec:discrepancy}

To look more closely at in what case the erroneous event detection happens, we show a chart in \cref{fig:max-error}, which represents the change in logarithmic luminance and the event occurrence at the pixel of the largest error. What is curious here is the curve of brightness change of our method almost follows that of the baseline method. This means that the estimate for the logarithmic luminance by our method is sufficiently accurate despite the early termination of path sampling. However, due to the spatiotemporal nature of event detection, one erroneous event detection can be an error source in subsequent event detection. Thus, as we can observe in \cref{fig:max-error}, the discrepancy of timings of event occurrence is small in the earlier part of the video, while the discrepancy gets larger in the latter part of the video. However, such error propagation might happen with real event cameras, as event cameras can suffer from many noise sources, including shot noise and electronic noise. Therefore, the error propagation shown here can be problematic only when comparing two sets of events and would not be a problem in practice.

\subsection{Limitations}
\label{ssec:limitaions}

\begin{figure}[t!]
  \centering
  \includegraphics[width=\linewidth]{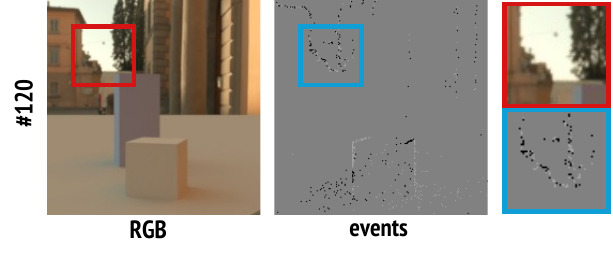}
  \caption{A failure case of the proposed method. At the region enclosed by a blue square of an event frame, failure events are detected even though the region enclosed by a red square in the RGB image, corresponding to the squared region in the event frame, is static.}
  \label{fig:failure}
\end{figure}

One notable limitation of our method is the failure event detection at the contour of static objects. \Cref{fig:failure} shows a typical example of the failure case. In this scene, the background scene is defined by an environment map, namely a high dynamic range image of the circumstance. Such background is typically far from the camera, and when the camera employs a thin-lens model or more realistic lens models, the depth of field blur often occurs there. As a result, the variance of brightness, particularly at the contour region, does not converge to a small value due to the blur. This problem may be resolved by considering the motion vectors as well in the rendering engine, and we would like to address it in future work.

In addition, we have assumed that the sample mean of the logarithmic luminance at each time point follows a normal distribution, which is evident from the central limit theorem. Currently, our experiments sample at most 4096 samples per pixel, which can be considered sufficient to satisfy the central limit theorem. However, when the sampling number is small, the sample mean may not follow a normal distribution, and the hypothesis test may not be valid. In such cases, we may need to first test whether the distribution of logarithmic luminance follows the normal distribution, not using another hypothesis testing method, such as the Shapiro--Wilk test and Kolmogorov--Smirnov test. Thus, it is a remaining challenge to consider the relationship between the number of samples and the distribution of the sample mean.



\section{Conclusion}
\label{sec:conclusion}

In this paper, we proposed the fully Monte-Carlo-based path tracing method for event rendering, where the event detection is performed statistically using hypothesis testing. As we have demonstrated with the experiments throughout this paper, the proposed method, while it significantly eliminates path tracing by discarding the null hypothesis, almost perfectly detects events that can be given by the baseline approach of simply tracing a full set of path samples. As a consequence, the computational time required by the proposed method is an order of magnitude less than that required by the simple baseline approach.

Since this study is the very first step to developing the Monte Carlo method for event rendering, there is much room for improvement to simulate the behavior of a real event camera, including current leakage, event threshold variation, motion blur, and shot noise, as addressed by the prior image-based event camera simulators~\cite{hu2021v2e, mcreynolds2023exploiting}. Despite the room, we believe implementing these behaviors in our current system will be a field of development rather than research, and we hope this very first study facilitates the development of event camera simulators based on Monte Carlo path tracing.

\ifpeerreview
\else
  \section*{Acknowledgments}
  This work was sponsored by the JSPS KAKENHI grant numbers JP24K02953, JP24K00783, and JST FOREST JPMJFR206I.

\fi

\ifbiblatex
  \printbibliography
\else
  \bibliographystyle{IEEEtran}
  \bibliography{references.bib}
\fi

\ifpeerreview
\else

\fi

\end{document}